\documentclass[11pt]{article}
\usepackage{coling2020}
\usepackage{times}
\usepackage{url}
\usepackage{latexsym}
\usepackage{tikz}
\usepackage{amsmath}
\usetikzlibrary{fit,positioning,arrows}
\usetikzlibrary{shapes,arrows}
\usetikzlibrary{intersections}

\tikzset{
  main/.style={circle, minimum size = 5mm, thick, draw=black!80, node distance = 10mm},
  connect/.style={-latex, thick},
  box/.style={rectangle, draw=black!100}
}

\usepackage{graphicx}

\title{MIDAS at SemEval-2020 Task 10: Emphasis Selection using Label Distribution Learning and Contextual Embeddings}

\author{Sarthak Anand\textsuperscript{1,3}\textsuperscript{*}, Pradyumna Gupta\textsuperscript{2}\textsuperscript{*}, Hemant Yadav\textsuperscript{3}\textsuperscript{*}, \\
\textbf{Debanjan Mahata\textsuperscript{4}, Rakesh Gosangi\textsuperscript{4}, Raymond Zhang\textsuperscript{4}, Rajiv Ratn Shah\textsuperscript{3}} \\ \\
\textsuperscript{1} NSIT-Delhi, India, \textsuperscript{2} IIT Dhanbad, India, \textsuperscript{3} IIIT Delhi, India, \textsuperscript{4} Bloomberg LP, New York \\
sarthaka.ic@nsit.net.in, pradyumna.17je003033@ece.iitism.ac.in, hemantya@iiitd.ac.in, \\ 
\{dmahata, rgosangi, hzhang449\}@bloomberg.net, rajivratn@iiitd.ac.in \\
}

\date{}

\begin{document}
\maketitle
\begin{abstract}
This paper presents our submission to the SemEval 2020 - Task 10 on emphasis selection in written text. We approach this emphasis selection problem as a sequence labeling task where we represent the underlying text with various contextual embedding models. We also employ label distribution learning to account for annotator disagreements. We experiment with the choice of model architectures, trainability of layers, and different contextual embeddings. Our best performing architecture is an ensemble of different models, which achieved an overall matching score of 0.783, placing us 15th out of 31 participating teams. Lastly, we analyze the results in terms of parts of speech tags, sentence lengths, and word ordering. \blfootnote{*Authors contributed equally.} 

\end{abstract}

\section{Introduction}
\blfootnote{This work is licensed under a Creative Commons Attribution 4.0 International Licence.} \blfootnote{Licence details: http://creativecommons.org/licenses/by/4.0/}
Emphasis selection is an emerging research problem \cite{shirani-etal-2019-learning} in the natural language processing domain, which involves automatic identification of words or phrases from a short text that would serve as good candidates for visual emphasis. This research is most relevant to visual media such as flyers, posters, ads, and motivational messages where certain words or phrases can be visually emphasized with the use of different color, font, or other typographic features. This type of emphasis can help with expressing an intent, providing more clarity, or drawing attention towards specific information in the text. Automatic emphasis selection is therefore useful in graphic design and presentation applications to assist users with appropriate choice of text layout. 

Prior works in speech processing \cite{mishra2012word,chen2017automatic} have modeled word-level emphasis using acoustic and prosodic features. Understanding emphasis in speech is critical to many downstream applications such as text-to-speech synthesis \cite{nakajima2014emphasized}, speech-to-speech translation \cite{do2015preserving}, and computer assisted pronunciation training \cite{felps2009foreign}. In computational linguistics, emphasis selection is very closely related to the problem of keyphrase extraction \cite{turney2002learning}. Keyphrases typically refer nouns and noun-phrases that capture the most salient topics in long documents such as scientific articles \cite{sahrawat2020keyphrase,mahata2018key2vec,swaminathan2020keyphrase}, news articles \cite{hulth2006study}, web pages \cite{yih2006finding}, etc. In contrast, emphasis selection deals with very short texts (e.g. social media posts), and also emphasis could be applied to words belonging to various parts of speech.

The goal of SemEval 2020 - Task 10 is to design methods for automatic emphasis selection in short texts. To this end, the organizers \cite{shirani2020semeval} provided a dataset consisting of over 3,000  sentences annotated for token-level emphasis by multiple annotators. The authors employed the standard I-O tagging schema, which is widely used in annotation of token-level tags. We approached emphasis selection as a sequence labeling task solved using a Bidirectional Long Short-term Memory (BiLSTM) model, where the individual tokens are represented using various contextual embedding models. We also employ label distribution learning (LDL) \cite{geng2016label} approach, which elegantly accounts for disagreements between the annotators. 

\section{Methods}

Let $d = \{w_1, w_2, ..., w_n\}$ be the input text, where $w_i$ is the $i^{th}$ token. The problem of emphasis selection is to assign each token $w_i$ one of two possible labels $E = \{e_I, e_O\}$, where $e_I$ denotes emphasis on the token and $e_O$ means otherwise. We approach this problem as a sequence labeling task solved using a BiLSTM model. We first represent each token $w_i$ with a dense vector $x_i$ of a fixed size. To this end, we explore three different embedding architectures: BERT \cite{devlin-etal-2019-bert}, RoBERTa \cite{DBLP:journals/corr/abs-1907-11692}, and XL-NET \cite{DBLP:journals/corr/abs-1906-08237}. Thus the given input text $d$ is transformed into a sequence of vectors $\{x_1, x_2, ..., x_n\}$. We then feed these vectors to a BiLSTM model which captures the sequential relations between the tokens. The hidden state of the BiLSTM $h_i$ is associated with the token $w_i$. Thus $h_i$ provides a fixed-size representation for token $w_i$ while incorporating information from the surrounding tokens. 

In standard sequence prediction problem, we can apply an affine transformation to map $h_i$ to the class space. However, in this paper, as with \cite{shirani-etal-2019-learning}, we employ LDL \cite{geng2016label} which transforms the output space into a distribution over the labels $E$. Namely, the objective of the model is not to just assign one label for a token but a real-valued vector. This vector is a distribution over the labels $E$, where the values are proportional to the number of annotations. To achieve this objective, we use KL-Divergence between the predictions and ground truth as the loss function for the model. 

\begin{equation}
\sum_{e_j \in E}p(e_j)\log{\frac{p(e_j)}{\tilde{p}(e_j)}}
\label{eqn:kldivergence}
\end{equation}

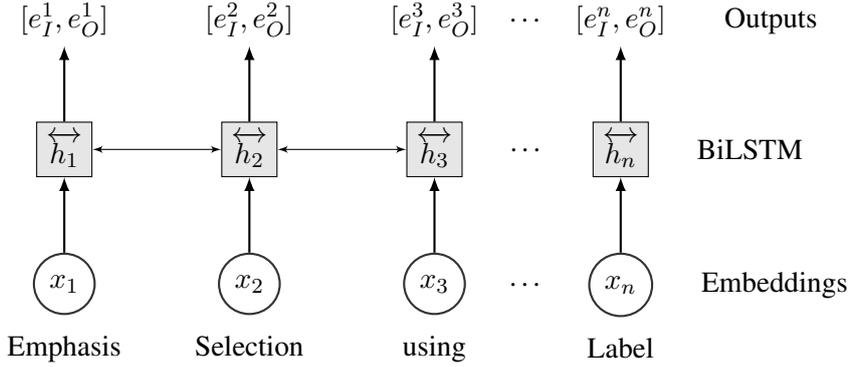
\begin{figure}
\centering
\begin{tikzpicture}
\node[text height=0.3cm] (y1) {$[e_I^1,e_O^1]$};
\node[text height=0.3cm] (y2) [right=of y1] {$[e_I^2,e_O^2]$};
\node[text height=0.3cm] (y3) [right=of y2] {$[e_I^3,e_O^3]$};
\node[text height=0.3cm] (yn) [right=of y3] {$[e_I^n,e_O^n]$};
\node[box,fill=black!10] (h1) [below=of y1] {$\overleftrightarrow{h_1}$};
\node[box,fill=black!10] (h2) [below=of y2] {$\overleftrightarrow{h_2}$};
\node[box,fill=black!10] (h3) [below=of y3] {$\overleftrightarrow{h_3}$};
\node[box,fill=black!10] (hn) [below=of yn] {$\overleftrightarrow{h_n}$};
\node[main] (x1) [below=of h1] {$x_1$};
\node[main] (x2) [below=of h2] {$x_2$};
\node[main] (x3) [below=of h3] {$x_3$};
\node[main] (xn) [below=of hn] {$x_n$};
\node[text width=2cm] (BiLSTM) [right=0.5cm of hn] {BiLSTM};
\node[text width=2cm] (Input) [right=0.5 of xn] {Embeddings};
\node[text width=2cm] (Output) [right=0.5 of yn] {Outputs};
\node[text height=0.3cm] (emphasis) [below=0.1cm of x1] {Emphasis};
\node[text height=0.3cm] (selection) [below=0.1cm of x2] {Selection};
\node[text height=0.3cm] (using) [below=0.1cm of x3] {using};
\node[text height=0.3cm] (label) [below=0.1cm of xn] {Label};
\path (y3) -- node[auto=false]{\ldots} (yn);
\draw[latex'-latex'] (h1) -- node[] {} (h2);
\draw[latex'-latex'] (h2) -- node[] {} (h3);
\path (h3) -- node[auto=false] {\ldots} (hn);
\path (h1) edge [connect] (y1);
\path (h2) edge [connect] (y2);
\path (h3) edge [connect] (y3);
\path (hn) edge [connect] (yn);
\path (x1) edge [connect] (h1);
\path (x2) edge [connect] (h2);
\path (x3) edge [connect] (h3);
\path (xn) edge [connect] (hn);
\path (x3) -- node[auto=false]{\ldots} (xn);
\end{tikzpicture}
\caption{BiLSTM architecture used for emphasis selection.}
\label{fig:architecture}
\end{figure}

The above equation demonstrates the loss for one sample, where $p(e_j)$ is the ground truth distribution and $\tilde{p}(e_j)$ is the model prediction. Note that the above equation reduces to negative log-likelihood in case of standard sequence prediction. The entire architecture is described in Figure \ref{fig:architecture}.

\section{Experimental work}

\subsection{Dataset}
The dataset provided for this SemEval task consists of 3,134 samples labeled for token-level emphasis by multiple annotators. The data was split into a training set consisting of 2,742 samples and development set with 392 samples. The training set has approximately 12 tokens per instance with the longest sample containing 38 tokens, and the shortest has one token. Likewise, the development set also has approximately 12 tokens per sample, and the longest sample has 31 tokens while the shortest has two tokens. Shirani et al. \cite{shirani-etal-2019-learning} has more details about the experimental protocols used for data collection. 

\subsection{Experimental settings}
We trained all the BiLSTM models using stochastic gradient descent in batched mode with the batch of 32. We used four different contextual embedding models for word representation: BERT (bert-base-uncased), BERT cased (bert-base-cased), RoBERTa (roberta-base), and XL-Net (xlnet-base-cased). We experimented on replacing the BiLSTM layer with a simple feed-forward dense layer. We also experimented with the trainabliity of different layers in the architecture. Namely, if none of the layers are trainable, only the last layer is trainable, and if all the layers are trainable. 

All the models were trained for 20 epochs: after each epoch, we evaluated on development dataset and stored the model from the best performing epoch. The hidden layers for the BiLSTM models were set to 128 units, the dense layers had 256 units, and the models trained at learning rates ranging from $2e-5$ to $3e-4$. We evaluated all the models in terms match scores as described in \cite{shirani-etal-2019-learning}. These match scores, for a given cardinality $m$, quantify the intersection between the top $m$ model predictions for emphasis and the ground truth as obtained from annotations.  

\section{Results}
\subsection{Architectures, Embeddings, and Trainablity}
Table \ref{tab:results} presents the performance of both the BiLSTM and the dense models for different choices of embeddings, and varying number of trainable layers. The first observation from these results is that the choice of architecture (BiLSTM vs. Dense) did not make a big difference in the performance. Second, the choice of embeddings did contribute significantly towards the performance: RoBERTa based models most often obtained the best scores, and XL-Net based models obtained the lowest scores. Lastly, we also observed that the model performance also improved with more trainable layers irrespective of the choice or architecture or embeddings. The best performing model, with an average match score of 0.788, was RoBERTa with a dense layer and all the layers set to be trainable. 

\begin{table}
\centering
\scalebox{0.8}{
    \centering
\begin{tabular}{|l|l |c|rrrr||r|}
\hline \bf Architecture & \bf Embedding & \bf Trainable layers & \bf M1 & \bf M2 & \bf M3 & \bf M4 & \bf Average\\ \hline
\hline
    BiLSTM & BERT &  & 0.592 & 0.739 & 0.797 & 0.833 & 0.740 \\
    & BERT Cased & None & 0.615 & 0.753 & 0.795 & 0.823 & 0.747 \\
    & RoBERTa &  & 0.587 & 0.745 & 0.804 & 0.836 & 0.743 \\
    & XL-Net & & 0.551 & 0.698 & 0.770 & 0.797 & 0.704 \\
\hline
\hline
    BiLSTM & BERT  &  & 0.602& 0.745& 0.801& 0.826& 0.744 \\
    & BERT Cased & Last Layer & 0.579& 0.758& 0.806& 0.832& 0.744 \\
    & RoBERTa & & 0.627& 0.752& 0.815& 0.845& 0.760 \\
    & XLNet & & 0.571& 0.710& 0.761& 0.811& 0.713\\
\hline
\hline
    BiLSTM & BERT  &  & 0.612& 0.752& 0.824 &0.840& 0.757 \\
    & BERT Cased & All & 0.610& 0.760& 0.805& 0.829& 0.750 \\
    & RoBERTa &  & 0.683& \bf 0.782& \bf 0.826 & \bf 0.853& 0.786 \\
    & XLNet &  &  0.584&  0.757&  0.799&  0.826& 0.742\\
\hline
\hline
    Dense & BERT &  & 0.589 & 0.733 & 0.792 & 0.827 & 0.735 \\
    & BERT Cased & None & 0.559 & 0.736 & 0.783 & 0.813 & 0.723\\
    & RoBERTa &  & 0.533 & 0.695 & 0.760 & 0.814 & 0.700 \\
    & XLNet & &  0.526 & 0.671 & 0.741& 0.786& 0.681\\
\hline
\hline
   Dense & BERT  &  & 0.582 & 0.744 & 0.801 & 0.832 & 0.740\\
   & BERT Cased & Last Layer &  0.586 & 0.750 & 0.798 & 0.8219 & 0.739\\
   & RoBERTa & & 0.630 & 0.761 & 0.810 & 0.836 & 0.759\\
   & XLNet & & 0.510 & 0.704 & 0.766 & 0.808 & 0.697 \\
   \hline
   \hline
   Dense & BERT  & & 0.630 &  0.776 & 0.810 & 0.837  & 0.763  \\
   & BERT Cased & All & 0.617 & 0.755 & 0.808 & 0.839 & 0.755  \\
   & RoBERTa & & \bf 0.702 &  0.776 & \bf 0.826 & 0.850 & \bf 0.788  \\
   & XLNet &  &0.602 & 0.745 & 0.813 & 0.847 & 0.752  \\
   \hline

\end{tabular}}
\caption{Performance of both model architectures (BiLSTM and Dense), for different choices of contextual embeddings and trainability of layers. The results are expressed in terms of match scores for four cardinalities, and the average of the four.}
\label{tab:results}
\end{table}

\subsection{Ensembling}
We experimented with two model ensembling approaches: average and weighted average. Average ensembling predicts the output simply as the average of outputs from all the models. In weighted averaging, we use the model performance on development dataset to weigh its contribution towards final prediction. We observed that the difference between these two ensembling approaches was rather minimal. We also tried ensembles of models with different combination of architectures and embeddings but eventually observed that the ensemble of all the models obtained the best performance. Table \ref{tab:ensemble} summarizes the results from some of these experiments. Our best system achieved an average match score of 0.783 on the final test dataset, placing us 15th out of 31 teams. The highest score achieved in the task was 0.823.

\begin{table}
\centering
\scalebox{0.8}{
    \centering
    \begin{tabular}{|c|c c c c|c| }
    \hline
       \bf Model & \bf M1 & \bf M2 &\bf M3 &\bf M4 & \bf Average  \\
    \hline
   RoBERTa (Dense \& BiLSTM) & 0.684 & 0.792 & 0.835 & 0.853 & 0.791 \\
   \hline
  RoBERTa, BERT, BERT Cased (Dense) & 0.658 & 0.801 & 0.842 & 0.862 & 0.791 \\
  \hline
  RoBERTa, BERT, BERT Cased (BiLSTM) & 0.679 & 0.790 & 0.835 & 0.859 & 0.791\\
 \hline
  RoBERTa, BERT, BERT Cased, XL-Net (Dense \& BiLSTM) & \bf 0.681 & \bf 0.801 & \bf 0.844 & \bf 0.864 & \bf 0.797\\
  \hline
    
    \end{tabular}}
    \caption{Performance of different ensembled models}
    \label{tab:ensemble}
\end{table}

\section{Analysis}

\subsection{Emphasis vs. Parts of Speech}
\begin{figure}[ht]
 \centering
 \scalebox{0.6}{
    \centering
     \includegraphics{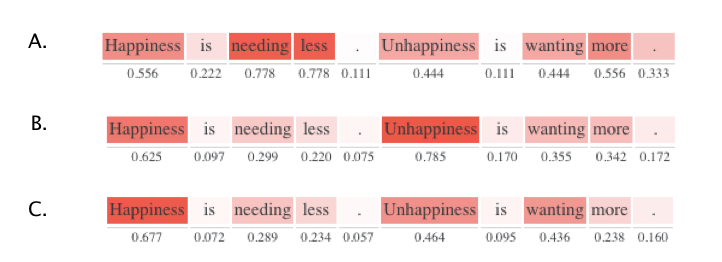}}
     \caption{Heat-map of: Ground truth (A), Roberta prediction (B), Bert prediction (C).}
     \label{fig:error_analysis_2}
 \end{figure}

We wanted to understand how the model predictions compared to the annotations for various parts of speech (POS) tags. Table \ref{tab:pos} presents the average emphasis score of human annotators on the development dataset for various POS tags. Also included in this table are the predictions from the best BERT and RoBERTa models. Of the various POS tags, nouns, proper-nouns, and adjectives are the classes with the most emphasis. This is also the case with the model outputs, however, they seem to be predicting higher emphasis scores for proper-nouns than nouns or adjectives. At the other end of the spectrum are coordinating conjunctions, adpositions, and punctuations. Figures ~\ref{fig:error_analysis_2} show an example of a situation where our models achieve very low match scores. Here, the models predicted the nouns 'Happiness' and 'Unhappiness' to have high emphasis but the annotators emphasized tokens which are verbs and adverbs. 

\begin{table}
    \centering
    \begin{tabular}{|l|c|c|c|c|}
            \hline 
            \bf Models  & \bf Count & \bf Humans & \bf BERT & \bf RoBERTa\\
               \hline
         Noun & 789&  0.528 & 0.463& 0.488 \\
         Verb & 855 & 0.305& 0.275& 0.295 \\
         Adjectives & 260 &   0.534& 0.436 & 0.482 \\
         ADP & 327& 0.136 & 0.139& 0.131 \\
         Det & 388 & 0.142& 0.133& 0.136\\
         Punctuations & 514 & 0.136& 0.141& 0.154 \\
         Adverbs & 263 & 0.290 &  0.249& 0.273 \\
         Pro-nouns & 325& 0.164&  0.154& 0.150 \\
         CCONJ & 88 & 0.130 &  0.143& 0.139 \\
         Proper-nouns & 156&   0.531& 0.504& 0.548 \\
         Part & 102& 0.212& 0.189& 0.195 \\
         Num & 32& 0.406& 0.365& 0.368\\
         X & 5 & 0.222& 0.262 & 0.298 \\
         INTJ & 3 & 0.481& 0.45 & 0.530 \\
         \hline
    \end{tabular}
    \caption{POS tags vs. average emphasis scores on dev dataset}
    \label{tab:pos}
\end{table}

\subsection{Shuffling Word Order}
We wanted to demonstrate that our models are not just picking up on certain keyphrases but capturing some important semantics in the data. To this end, we trained a new set of models on the training dataset, where for each sample, the order of the words was randomly shuffled. The resulting models were then evaluated on the development dataset. We repeated this experiment five times and Table \ref{tab:shuffle} presents the average performance across these runs. Also included in this table is a baseline model which predicts a random score for each token. As expected, the models trained on shuffled data are significantly worse than their counterparts in Table \ref{tab:results}. Another interesting observation is that the performance of these models is comparable to the random baseline. This suggests that the word order and therefore semantic structure is very important to the emphasis selection problem.

\begin{table}
    \centering
    \begin{tabular}{|l|l|rrrr||c|}
\hline \bf Architecture & \bf Embedding  & \bf M1 & \bf M2 & \bf M3 & \bf M4 & \bf Average\\ \hline
      BiLSTM & BERT & 0.179&0.297&0.382&0.453&0.328 \\
      BiLSTM & BERT Cased  &  0.168&0.280&0.368&0.438  &0.314 \\
      BiLSTM & RoBERTa  &0.173&0.292&0.378&0.446&0.322 \\
      BiLSTM & XL-Net  &0.171&0.286&0.377&0.448&0.321 \\
      Dense & BERT   & 0.171&0.294&0.377&0.441&0.321 \\
      Dense & BERT Cased  &  0.175&0.289&0.368&0.437  &0.317 \\
      Dense & RoBERTa  &0.178&0.282&0.374&0.444&0.319 \\
      Dense & XL-Net  &0.157&0.280&0.365&0.444&0.311 \\
      \hline
      \multicolumn{2}{|c|}{Random} & 0.175 & 0.276 & 0.352 & 0.428 & 0.308 \\
      \hline
    \end{tabular}
    \caption{Performance of models trained on data where the sentences were randomly shuffled.}
    \label{tab:shuffle}
\end{table}

\subsection{Length vs. Performance}
We also wanted to understand how the model performance is influenced by the length of the samples. As mentioned earlier the average length of each sample in the dataset is 12 tokens and the standard deviation of the length is around 6. Driven by these statistics, we decided to split the development data into three sets: \textit{Short} ($<$ 6 tokens, 80 samples), \textit{Medium} (6 to 18 tokens, 262 samples), and \textit{Long} ($>$18 tokens, 50 samples). Table \ref{tab:length} summarizes the results (average match score) of all the models split into these three groups. All the models, irrespective of the choice of architecture or embeddings, have deteriorated with increasing length of the samples. The difference between the longest and shortest samples is most pronounced for BERT-based models. RoBERTa-based models seem to be handling longer samples much better than the other two embeddings.

\begin{table}
    \centering
    \begin{tabular}{|l|l|c c c|}
    \hline
        \bf Architecture & \bf Embedding & \bf Short & \bf Medium & \bf Long   \\
        \hline
        Dense & BERT & 0.850 & 0.776 & 0.664 \\
        BiLSTM & BERT & 0.870 & 0.762 & 0.656 \\
        Dense & BERT Cased & 0.850 & 0.762 & 0.668 \\
        BiLSTM & BERT Cased & 0.852 & 0.763 & 0.637 \\
        Dense & RoBERTa & 0.863 & 0.801 & 0.702 \\
        BiLSTM & RoBERTa & 0.873 & 0.793 & \bf 0.710 \\
        \hline
        \multicolumn{2}{|c|}{Best Ensemble} & \bf 0.875 & \bf 0.808 & 0.709 \\
    \hline
    \end{tabular}
    \caption{Average match score vs. length of the sample}
    \label{tab:length}
\end{table}

\section{Conclusion}
In this paper, we present our submission to the SemEval 2020 - Task 10 on emphasis selection in written text. Our best performing model achieved an overall matching score of 0.783, placing us 15th out of 31 participating teams. We approached emphasis selection as sequence prediction problem solved using BiLSTMs. Our experimental work demonstrates the effect of model architectures, trainability of layers, and embeddings on the performance. We analyze the results in terms of parts of speech tags and sentence lengths. Our analysis provides some interesting insight into some of the shortcomings of the models and also the challenges with emphasis selection. 

\bibliographystyle{acl}
\bibliography{coling2020}

\end{document}